\documentclass[lettersize,journal]{IEEEtran}
\usepackage{mathptmx}
\usepackage{amsmath,amsfonts}
\usepackage{algorithm}
\usepackage{algorithmic}
\usepackage{array}
\usepackage[caption=false,font=normalsize,labelfont=sf,textfont=sf]{subfig}
\usepackage{textcomp}
\usepackage{stfloats}
\usepackage{url}
\usepackage{verbatim}
\usepackage{graphicx}
\usepackage{pifont}

\usepackage{multirow}
\usepackage{multicol}

\usepackage{xcolor,colortbl}
\usepackage{booktabs}       
\usepackage{amsfonts}

 \DeclareMathAlphabet{\mathcal}{OMS}{cmsy}{m}{n}
 
\usepackage{amssymb}
\usepackage{mathtools}

\usepackage{siunitx}
\usepackage{xspace}
\usepackage{lipsum}

\usepackage{adjustbox}
\newcommand{\e}{\epsilon}
\newcommand{\y}{\gamma}

\newcommand{\E}{\mathbb{E}}
\newcommand{\N}{\mathcal{N}}

\newcommand{\B}{\mathcal{B}}

\DeclareMathOperator*{\argmin}{argmin}
\DeclareMathOperator*{\argmax}{argmax}

\def\BibTeX{{\rm B\kern-.05em{\sc i\kern-.025em b}\kern-.08em
    T\kern-.1667em\lower.7ex\hbox{E}\kern-.125emX}}
\usepackage{balance}
\IEEEoverridecommandlockouts

\title{Improving Policy Exploitation in Online Reinforcement Learning with Instant Retrospect Action}

\author{Gong Gao$^{1}$, Weidong Zhao$^{1*}$,
Xianhui Liu$^{1}$, and Ning Jia$^{1}$
\thanks{$^{1}$ School of Computer Science, Tongji University, China 
{\tt\small g18438613630@126.com},
}%
}

\begin{document}
\maketitle
\thispagestyle{empty}
\pagestyle{empty}
\begin{abstract}
Existing value-based online reinforcement learning (RL) algorithms suffer from slow policy exploitation due to ineffective exploration and delayed policy updates.
To address these challenges, we propose an algorithm called Instant Retrospect Action (IRA). Specifically, we propose Q-Representation Discrepancy Evolution (RDE) to facilitate Q-network representation learning, enabling discriminative representations for neighboring state-action pairs. In addition, we adopt an explicit method to policy constraints by enabling Greedy Action Guidance (GAG). This is achieved through backtracking historical actions, which effectively enhances the policy update process. Our proposed method relies on providing the learning algorithm with accurate $k$-nearest-neighbor action value estimates and learning to design a fast-adaptable policy through policy constraints. We further propose the Instant Policy Update (IPU) mechanism, which enhances policy exploitation by systematically increasing the frequency of policy updates. We further discover that the early-stage training conservatism of the IRA method can alleviate the overestimation bias problem in value-based RL. Experimental results show that IRA can significantly improve the learning efficiency and final performance of online RL algorithms on eight MuJoCo continuous control tasks. The code is available at https://github.com/2706853499/IRA.
\end{abstract} 
\begin{IEEEkeywords}
    online reinforcement learning, slow policy exploitation, representation learning, policy constraints
    \end{IEEEkeywords}

\IEEEpeerreviewmaketitle
\section{Introduction}
\label{Introduction}
Online reinforcement learning (RL) is a branch of Artificial Intelligence (AI) that studies how agents make optimal decisions in complex dynamic environments. This decision-making process relies on the agents' actions and the feedback signals they receive from the physical environment. The combination of deep neural networks and RL has achieved increasing success in real-world RL tasks such as gaming AI~\cite{wang2024negatively,shaheen2025reinforcement}, robot control~\cite{qi2023adaptive,radosavovic2024real,zhuang2025tdmpbc}, and medical diagnosis~\cite{liu2024segmenting,wang2024deep,luo2024dtr}. 
Nonetheless, a persistent challenge in value-based online RL arises from slow policy exploitation~\cite{quangaugmenting}, which is attributed to delayed policy update techniques and epistemic uncertainties in Q-value estimation. 
Specifically, these issues pertain to the agent's inability to rapidly adapt to dynamic environmental changes, leading to suboptimal performance of the learned policies.

Based on the algorithms dedicated to enhancing policy exploitation, existing RL research can generally be divided into two categories:
i) Representation learning algorithms, which aim to extract more effective feature representations to enhance the decision-making process. They achieved this by either training auxiliary networks~\cite{yarats2021improving,ni2023transformers,fujimoto2024sale,ni2024bridging,quangaugmenting} or introducing additional supervisory signals~\cite{he2022reinforcement,he2023frustratingly,zheng2024effective}. 
However, these methods leverage auxiliary networks to reconstruct historical interaction data to enhance state perception, which inherently violates the Markov property~\cite{allen2021learning} of decision-making processes and incurs significant memory overhead.
Moreover, integrating additional supervisory mechanisms demands careful alignment between the design of auxiliary loss functions and RL frameworks. This alignment is crucial to mitigate gradient conflicts and preserve the stability of the learned policy.
ii) Policy constraint algorithms~\cite{shah2018q,ma2023iteratively,tarasov2024revisiting}, which constrain policies to avoid ineffectively exploring high uncertainty action space. These algorithms integrated behavior cloning~\cite{fujimoto2021minimalist,ran2023policy} into the Actor-Critic framework~\cite{cheng2022adversarially}, thereby enabling the policy to maintain a pessimistic nature while retaining a certain level of exploration capability. 
Despite the success of policy constraint and its extensions, previous algorithms combining behavior cloning with RL have mainly focused on offline RL~\cite{liu2024adaptive,mao2025offline}. 
However, offline RL and its pessimistic properties are not applicable in the field of online RL.

\begin{figure}[!ht]
\centering
\includegraphics[width=0.5\textwidth]{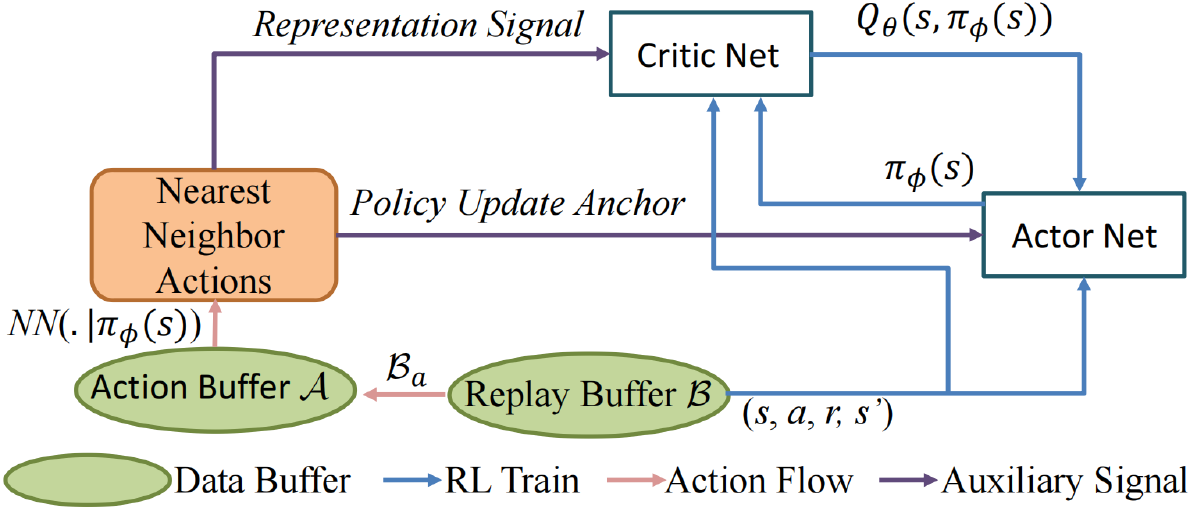}
\caption{We introduce auxiliary signals to enhance learning capability and propose two core mechanisms: integrating representation-guided signals into Q-learning and introducing anchor points for policy updates.}
\label{overall_architecture}
\end{figure}
To address the aforementioned limitations, we propose an Instant Retrospect Action (IRA) algorithm, which consists of three modules: i) Introducing a Q-representation signal aiming to improve the representational discriminability of neighboring actions; ii) Employing a policy update anchor to amplify policy exploitation, while preserving constrained exploration capacity through bounded policy exploration; iii) Enabling instant policy updates to enhance policy exploitation (see Section~\ref{sec:method}).
The proposed method serves as an augmenting component integrated into a standard value-based algorithm, preserving the intrinsic advantages of value-based learning. We instantiate the method with practical implementations based on TD3\cite{fujimoto2018addressing} and DDPG~\cite{lillicrap2015continuous}, which effectively enhance policy exploitation and lead to improved performance.

The contributions of this work are fourfold:

\begin{itemize}

\item  We propose an IRA algorithm that employs an explored action buffer as an explicit constraint for policy updates, effectively mitigating the slow exploitation dilemma in value-based RL. 
\item  Our proposed method introduces representation discrepancy between neighbor actions in given states as auxiliary supervisory signals, enabling the Q-network to learn value-discriminative embeddings at the feature level. 

\item By introducing optimal actions as explicit policy guidance, the proposed method effectively avoids inefficient exploration patterns. Meanwhile, the conservatism in this guided exploration helps mitigate the overestimation bias. 
\item We implement an instant policy update mechanism to further boost policy exploitation. 
Extensive evaluations on continuous control benchmarks demonstrate 36.9\% average performance gain over vanilla TD3 across MuJoCo tasks. 
\end{itemize}

The remainder of this paper is organized as follows. Section~\ref{sec:pre} introduces the background. Related work is reviewed in Section~\ref{related_wor}. In Section~\ref{sec:method}, we present the proposed Instant Retrospect Action (IRA), including the overall framework and implementation details. Section~\ref{sec:experiment} reports experimental results on 8 challenging continuous control tasks, demonstrating the effectiveness and practicality of IRA. Finally, Section~\ref{conclusion} concludes the paper and outlines potential future directions.

\section{Preliminaries}
\label{sec:pre}

\textbf{Value-based RL.}
In this paper, we consider a $(S, A, T, R, \gamma)$ by-definition deterministic Markov Decision Process (MDP)~\cite{kumar2023policy} in which RL is carried out, where $S$ is the state space, $A$ is the action space, and $\gamma \in [0,1)$ is the discount reward factor. The environment is characterized by two attributes: the transition function $T:S \times {A} \to {S}$, and the reward function $R:S \times {A} \to {R}$. The output of RL is the policy $\pi :S \to A$, which maximizes the objective of RL $J(\pi )= {\E_s}[\sum\nolimits_{t = 0}^\infty  {{\gamma ^t}{r_{t}}|{a_t} = } \pi ({s_t})]$, where ${\E}$ denotes the expectation.

A typical value-based RL algorithm can be defined by integrating the two Q-learning optimization processes. The first involves training the Q-network using temporal difference error, which can be formulated as
\begin{equation} 
\label{eq:belleman}
\theta^{*} =\argmin_{\theta}\E_{s,a,{s^{'}},r}{[Q_\theta(s,a )-y]^2}, \\
\end{equation}
where $y=r + \y \max_{a^{'}} Q_{\theta'}({s^{'}}, a^{'})$ is the Bellman optimal Q-value, $s, a, s^{'}, a^{'}, r$ is state, action, next state, next action and immediate reward, respectively. The learning process of the Actor-network can be formalized as
\begin{equation}
{\phi^{*}} = {\mathop {\argmax}\limits_{\phi}}{\E_{s}}[{Q_\theta }(s,\pi_{\phi}(s))].
\end{equation}

The goals of the training process include obtaining high-quality value estimators and then using these estimators to guide policy improvement. This algorithm (such as TD3~\cite{fujimoto2018addressing}) currently achieves state-of-the-art results on a wide range of task sets.
The TD3 algorithm employs techniques such as double Q-learning~\cite{van2016deep} and Q-target networks~\cite{mnih2013playing}, which progressively stabilize the estimates of the Q-value. 
However, value-based RL algorithms suffer from slow exploitation, leading to suboptimal policies.

\textbf{Q Representation.}
Following the Q-network representation paradigm defined in PEER~\cite{he2023frustratingly}, we decompose the network parameters $\theta $ into two distinct components: ${\theta _ + }$ and ${\theta _{ - }}$. The Q-value computation can be formulated as
\begin{equation}
    \centering
    Q(s,a;\theta ) =  < \varphi (s,a;{\theta _ + }),{\theta _{ - }} > ,
\end{equation}
where, $\theta_+$ represents the nonlinear encoder, while $\theta_-$ denotes the parameters of the linear projection used to estimate the Q-value. Specifically, the learned representation is explicitly defined as the output of the encoder $\theta_+$ through the function $\varphi$, establishing a hierarchical decomposition of feature abstraction and value estimation.

\textbf{Policy Constraint.}
Policy constraints are routinely applied in offline RL to ensure actions remain consistent with the static dataset, and are widely adopted due to their broad applicability and effectiveness. 
This algorithm has become increasingly crucial in offline RL and offline-to-online RL ~\cite{yu2023actor,zheng2023adaptive,luo2025optimistic}, accelerating policy exploitation through constrained policy exploration.
In this approach, a constraint policy is applied to ensure that the policy ${\pi _\phi }$ remains proximate to the trained action set ${\mathcal{A}}$, which can be formulated as
\begin{equation}
 ||{\pi _\phi }(s)- {NN}(.|{\pi _\phi(s) })  || \le \varepsilon_a,
\end{equation}
where $||.||$ indicates Euclidean distance, $\varepsilon_a $ denotes a constant value that the divergence between the policy ${\pi _\phi }$ and the nearest neighbor action ${NN}(.|{\pi _\phi(s) })$ is bounded. 
This approach ensures that policy updates remain bounded, thereby improving the stability of the policy.

\section{Related Works}
\label{related_wor}
\textbf{Online RL.}
In general, there are three primary algorithms for solving online RL problems: value-based online RL, policy-based online RL, and representation learning-based online RL.  
In value-based online RL, value estimators guide policy improvement by providing reliable gradient directions, as emphasized in prior work~\cite{lillicrap2015continuous,fujimoto2018addressing}.
In policy-based online RL, algorithms such as Proximal Policy Optimization (PPO)~\cite{schulman2017proximal} indirectly optimize the target policy through advantage estimation.
On the other hand, representation learning-based online RL leverages representation learning algorithm to improve the approximation of the policy function, such as~\cite{quangaugmenting}.

\textbf{Representation Learning-based RL.}
Many existing RL algorithms enhance the efficiency of representation learning by introducing auxiliary networks~\cite{ni2023transformers,fujimoto2024sale,ni2024bridging,quangaugmenting,amortila2025reinforcement} or additional supervisory signals~\cite{he2022reinforcement,he2023frustratingly,zheng2024effective}. 
On one hand, several scholars~\cite{ni2023transformers,fujimoto2024sale,ni2024bridging,quangaugmenting,amortila2025reinforcement} have investigated the simultaneous utilization of observed states and historical state representations to augment the decision-making process, thereby enhancing sampling efficiency in observation-rich environments. 
However, these algorithms experienced error accumulation due to reliance on historical representations for decision enhancement, ultimately leading to suboptimal policy.

On the other hand, several scholars~\cite{he2022reinforcement,he2023frustratingly,zheng2024effective} proposed leveraging additional loss functions to better learn representations. These algorithms applied to value-based~\cite{lillicrap2015continuous,fujimoto2018addressing} RL.
PEER~\cite{he2023frustratingly} utilized the differences between temporally adjacent state-action pairs as an additional supervisory signal in RL training to enhance representational disparity, thereby implicitly improving the exploration capability.
Our approach shares similarities with PEER in leveraging additional supervisory signals for representation learning. However, the key distinction lies in our approach's focus on addressing the challenge of accurately estimating $k$-nearest neighbor action values within the context of learning state-action embeddings.
The RDE approach aims to provide optimal value actions for policy updating by our proposed GAG approach. 

Our core innovation lies in guiding policy representations away from suboptimal action embeddings, while explicitly constraining the policy updates to approach locally optimal actions.

\textbf{Implicit and Explicit Constraints.}
Existing implicit~\cite{schulman2017proximal} and explicit~\cite{fujimoto2021minimalist,shen2021theoretically,chemingui2024offline,fan2025solve} policy constraint algorithms in RL enhance policy improvement by incorporating a clipping function or a penalty term into the Actor-network.
Proximal Policy Optimization (PPO)~\cite{schulman2017proximal} employed a clipped function to constrain the magnitude of policy updates, thereby preventing drastic changes within a single iteration and reducing the risk of training instability.
Since the clipping mechanism in PPO does not explicitly incorporate prior knowledge, it may suffer from slow convergence due to insufficient guidance. 
In contrast, explicit constraint algorithms such as behavior cloning can directly impose policy constraints using expert data, which helps alleviate the issue of slow policy exploitation.
In the field of offline RL, many scholars employed behavior cloning to implement conservative policy constraints.

Similar to explicit policy constraint algorithms~\cite{chemingui2024offline,fan2025solve}, which imposed policy constraints by identifying optimal global actions within the action space. Our proposed IRA is strikingly different from the aforementioned works as they retrieve exhaustively over the entire global action space. Due to their high computational complexity, these approaches are primarily applicable to discrete action spaces. 
Furthermore, NNAC~\cite{shen2021theoretically} incorporated a nearest-neighbor state-action constraint to facilitate temporal-difference learning and policy updates in continuous action spaces.
However, this algorithm imposes strict constraints on policy exploration, resulting in overly conservative behavior, which may ultimately lead to the learning of suboptimal policies. 

In contrast, our approach employs the top-$k$ nearest optimal actions as explicit constraints, striking a balance between policy exploration and improving policy exploitation.

\section{Methods}
\label{sec:method}
In conjunction with value-based RL algorithms, as illustrated in Algorithm~\ref{alg:IRA}, we first retrieve the $k$-nearest neighbors of the current policy and utilize the Q-target network to differentiate between optimal and suboptimal actions. Subsequently, we introduce a Q-representation discrepancy evaluation loss to enhance the representational disparity between the predicted action and suboptimal actions.
To achieve constrained policy improvement through systematic exploration, we impose explicit policy constraints via greedy action guidance, ensuring proximity to the optimal actions.
Finally, we propose an instant action update mechanism to further enhance policy exploitation. 

\begin{algorithm}[tb]
   \caption{Instant Retrospect Action}
   \label{alg:IRA}
\begin{algorithmic}
   \STATE Initialize critic networks $Q_{\theta_{1}}$, $Q_{\theta_{2}}$, and actor-network $\pi_\phi$ with random parameters $\theta_{1}$, $\theta_{2}$, $\phi$
   \STATE Initialize target networks $\theta'_1 \leftarrow \theta_1$, $\theta'_2 \leftarrow \theta_2$, $\phi' \leftarrow \phi$
   \STATE Initialize replay buffer $\B$
   \STATE Initialize explored action buffer ${\mathcal{A}}$, size $n$
   \FOR{$t=1$ {\bfseries to} $T$}
   \STATE Select action with exploration noise $a \sim \pi_\phi(s) + \e$, 
   \STATE $\e \sim \N(0, \sigma)$ and observe reward $r$ and new state $s'$
   \STATE Store transition tuple $(s, a, r, s')$ in $\B$
    \STATE Store action data $a$ in $\mathcal{A}$
   \STATE Sample mini-batch of transitions $(s, a, r, s')$ from $\B$
   \STATE Find the set of actions ${\hat {\mathcal{A}}} \in {\mathcal{A}}$ that are closest to the policy $\pi_\phi$
   \STATE Find the optimal action ${{\tilde a}_{opt}} \in \hat {\mathcal{A}}$ and the suboptimal action ${{\tilde a}_{sub}} \in \hat {\mathcal{A}}$ 
   \STATE Update critic by minimizing Eq.~\eqref{eq:critics}
   \IF{$t$ mod $d$}
   \STATE Update actor by maximizing Eq.~\eqref{eq:pi}
   \STATE $\theta'_i \leftarrow \tau \theta{ _i} + (1 - \tau) \theta'_i$
   \STATE $\phi' \leftarrow \tau \phi + (1 - \tau) \phi'$
   \ENDIF
   \ENDFOR
\end{algorithmic}

\end{algorithm}

We employed the Chebyshev distance~\cite{deza2009encyclopedia} to calculate the distance between the current predicted action and the actions in the action buffer ${\mathcal{A}}$, subsequently ranking ${\mathcal{A}}$. This can be formulated as
\begin{equation}
\label{eq:_sort}
{{{\mathcal{A}}}^{'} } = {\rm{sort}_{\textit{a} \in {\mathcal{A}},\rm{ascending}}} \max(|\pi_\phi (s)_j - a_j|),
\end{equation}
where ${\rm{sort}_{\rm{ascending}}}$ denotes the operation of sorting all elements $a \in {\mathcal{A}}$ in ascending order, $\rm{max}(|.|)$ denotes Chebyshev distance, and $j$ indicates the index of the action dimension. We can then identify the $k$-nearest actions ${{\mathcal{A}}}^{'}$ that are closest to the action predicted by the policy $\pi_\phi(s)$, which can be formulated as
\begin{equation}
\label{eq:_nearest_n_2}
\hat {{\mathcal{A}}}=\mathop {{\rm{top}}}\limits_k  { {{\mathcal{A}}}^{'}},
\end{equation}
where, $\mathop {\text{top}}\limits_k$ denotes the selection of the top $k$ items from ${{\mathcal{A}}}^{'}$.
We leverage double Q-target networks to obtain the true Q-values for each state-action pair in descending order, which can be formulated as
\begin{equation}
\tilde {{\mathcal{A}}}  = {\rm{sort}_{{a \in \hat{A}},\rm{descending}}} \min_{i=1,2}( Q_{\theta'_i}(s, {a})) ,
\end{equation}
where ${\tilde {{\mathcal{A}}} }$ denotes the ordered set of actions, ${\rm{sort}_{\rm{descending}}}$ denotes the operation of sorting all elements $a \in \hat {\mathcal{A}}$ in descending order. 
Subsequently, we identify the optimal action $\tilde{a}_{opt}={\tilde {{\mathcal{A}}}}[0]$ and the suboptimal action $\tilde{a}_{sub}={ \tilde {{\mathcal{A}}} }[1]$ of the action buffer ${\mathcal{A}}$.
Upon identifying the optimal and suboptimal actions, an intuitive optimization approach is to amplify the representational divergence between the current and suboptimal policies in the state-action space. Concurrently, the policy network undergoes progressive optimization by aligning policy update anchors with the optimal actions through a supervised learning paradigm.

\subsection{Representation Discrepancy Evolution}
We introduced a representation regularization term that specifically increases the disparity for the nearest neighbor actions in the current state. 
According to the Q-network representation established in Section \ref{sec:pre}, the Q-network's representation is formulated as $\varphi (s, a; \theta_{+})$.

In value-based RL, the Q-network assesses the quality of an action taken by an agent within a given state by computing the action value function. 
To enhance the representational difference between current policy predictions and suboptimal actions, and to provide an explicit high-value action for policy improvement, we employ a straightforward representational regularization technique.
This algorithm amplifies differentiation in Q-representations, thereby implicitly enhancing action distinctions, ultimately leading to more effective greedy action guidance.

The similarity between $\varphi(s, \pi_{\phi}(s); \theta_+)$ and $\varphi(s, \tilde{a}_{\text{sub}}; \theta_+^{\prime})$, the state-action representations produced by the Q-network, is defined as their inner product $< \cdot, \cdot >$, and can be formulated as
\begin{equation}
  < \varphi (s,\pi_{\phi}(s);{\theta _ + }),\varphi (s,{\tilde {a}_{sub}};{\theta^{'} _+}) >  ,
\end{equation}
where $s$, $\pi_{\phi}$, ${\tilde {a}_{sub}}$, ${\theta _ + }$, and ${\theta^{'} _ + }$ denote the state, the Actor-network, the suboptimal action under state $s$, the parameters of the Q-network except for the last fully connected layer, and the parameters of the Q-target network except for the last fully connected layer, respectively.
In particular, Q-representation discrepancy evolutionary loss can be defined as
\begin{equation}
{L_{RDE}}(\theta ) =\alpha  < \varphi (s,\pi_{\phi}(s);{\theta _ + }),\varphi (s,{\tilde {a}_{sub}};{\theta^{'} _ + }) > ,
\end{equation}
where $\alpha$ denotes the intensity parameter of the RDE loss.
In this setup, the Q-network takes the state $s$ and the action $\pi_{\phi}(s)$ as inputs, while the Q-target network utilizes the state $s$ and the suboptimal action ${\tilde{a}_{sub}}$ as inputs.
The goal is to widen the gap between the two actions in the representation space to contribute to the assessment of the value of the action made by the Q-network. The loss of the Q-network consists of the temporal difference error and the RDE loss, which can be formulated as
\begin{equation}
\label{eq:critics}
\centering
   {{L}_Q}(\theta ) = {[Q_{\theta}(s,a ) - y]^2} + {L_{RDE}}(\theta ) ,
\end{equation}
where $y=R(s,a) + \y \min_{i=1,2} Q_{\theta'_i}(s',{\pi _{\phi^{'}} }(s^{'}))$.
A smaller RDE loss indicates a greater representational difference between the predicted action $\pi _{\phi} (s)$ and the suboptimal action ${\tilde {a}_{sub}}$. Conversely, a larger RDE loss suggests a smaller representational difference between them, which may cause the agent to get trapped in locally optimal regions, resulting in suboptimal performance.
However, due to the problem of overestimation of Q-networks due to epistemic uncertainty during the RL training phase,  which hampers effective exploration, we propose a greedy action guidance to address this.

\subsection{Greedy Action Guidance}
\label{sec:mag}
After determining the optimal nearest neighbor $\tilde{a}_{opt} \in {\mathcal{A}}$ for the predicted action $\pi (s)$, our goal is for the current actor's decisions to align closely with high-value actions stored in the buffer ${\mathcal{A}}$. To achieve this, we impose explicit policy constraints on the current actor-network to encourage effective policy exploitation.

By treating the constraint term as a penalty, we maximize the following objective:
\begin{equation}
\label{eq:KLreg}
\max_\phi \mathbb{E}_{s} [Q_\theta(s,{\pi _\phi }(s) ) -  || \pi_\phi(s)-{NN}(.|\pi _\phi(s))||],
\end{equation}
where ${NN}(.|\pi _\phi(s)$ denotes the process that identifies neighboring optimal actions from the current policy's predictions $\pi_\phi(s)$, enforcing explicit policy constraints through localized optimization in the action space.
With optimal action $\tilde a_{opt}$ defined, Eq.~\eqref{eq:KLreg} is equivalent to the following policy training objective:
\begin{equation}
\label{eq:pi}
\begin{aligned}
J_{\pi}(\phi)=   {\E_{s}}\left[ {{(Q_\theta }(s,{\pi _\phi }(s) - \mu {{({\pi _\phi }(s) - {{\tilde a}_{opt}})}))^2}} \right],
\end{aligned}
\end{equation}
where $\mu$ denotes the policy constraint strength and $ \mu {{\left({\pi _\phi }(s) - {{\tilde a}_{opt}}\right)}^2}$ computes the overall constraint.
By Eq.~\eqref{eq:pi}, ${\pi_\phi}$ explores within this constrained exploration region, thereby facilitating gradual policy improvement.
However, a fixed $\mu$ may lead to conservative policy due to insufficient exploration. Therefore, we implement a stepwise decay of $\mu$ in our RL training, decreasing it to 0.1. This ensures that greedy action guidance continuously guides policy update throughout the training phase, leading to performance enhancement.

\subsection{Instant Policy Update}
In distributed reinforcement learning~\cite{yu2024cheaper}, dynamically adjusting policy update frequency or step size is a widely adopted technique to enhance policy utilization efficiency. This mechanism has been extensively validated in frameworks such as Asynchronous Advantage Actor-Critic (A3C)~\cite{mnih2016asynchronous} and the Importance Weighted Actor-Learner Architecture (IMPALA)~\cite{espeholt2018impala}, where adaptive control of parameter synchronization intervals demonstrates significant improvements in both sample efficiency and computational resource allocation.
We argue that a delayed policy update ($d$=2) hinders the instant adjustment of the policy, further exacerbating the slow exploitation phenomenon. To address this, we propose an instant policy update ($d$=1) mechanism, which increases the update frequency of the actor-network, effectively mitigating the slow exploitation issue.

\section{Experiments}
\label{sec:experiment}

In this section, we employ eight online MuJoCo continuous tasks to evaluate the performance of the IRA, aiming to answer the following questions: i) How does IRA perform in various continuous control tasks compared to previous algorithms? ii) Can IRA address the slow exploitation problem in value-based RL algorithms?
iii) Is IRA superior to other representation learning and policy constraint algorithms?
\subsection{Setup for Empirical Evaluation}
\begin{figure*}[!htbp]
\begin{center}
\centerline{\includegraphics[width=0.7\textwidth]{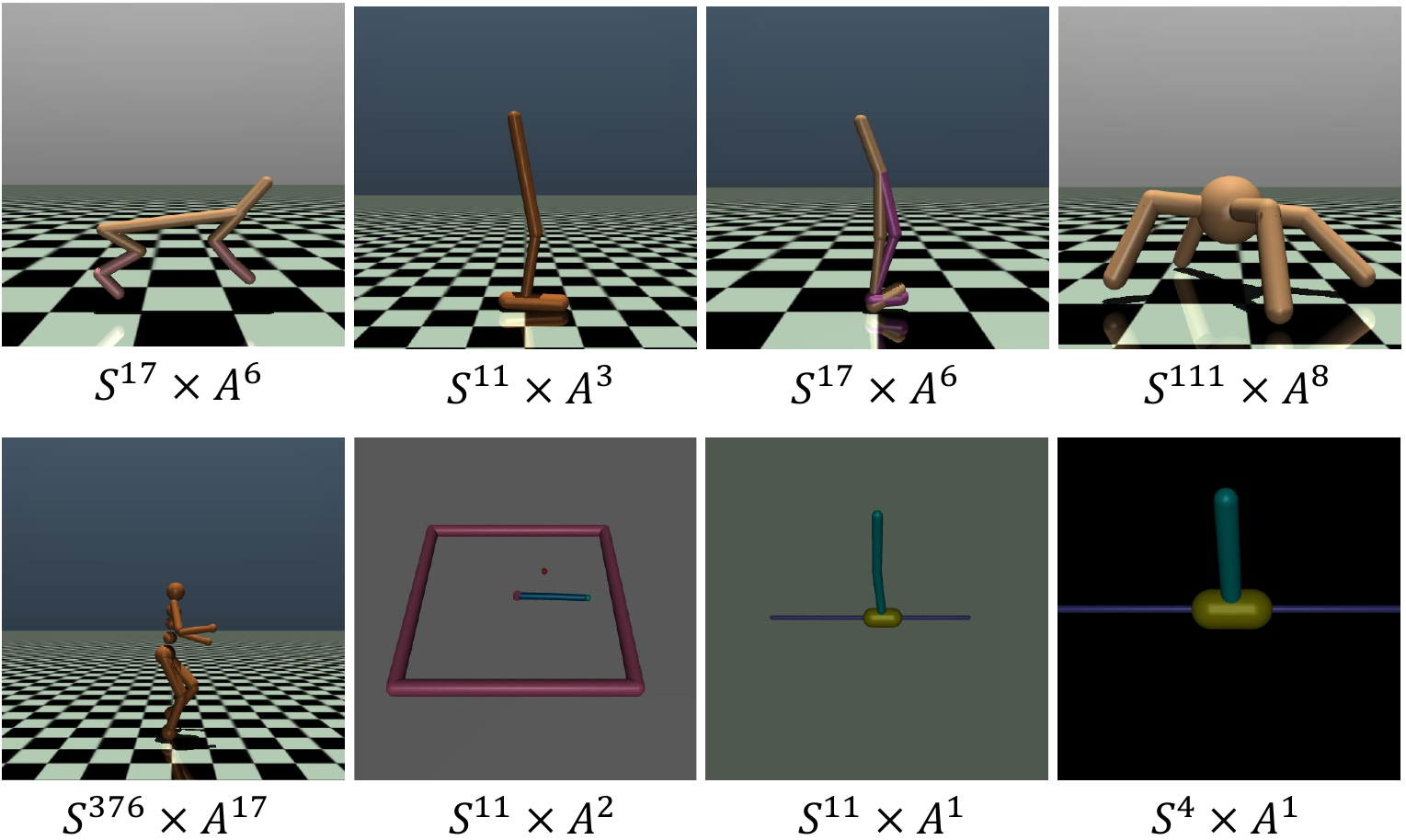}}
\caption{Images for eight MuJoCo environments used in our experiments.}
\label{task_visual}
\end{center}
\end{figure*}

\textbf{MuJoCo Tasks.}
We evaluate the effectiveness of IRA across eight standard MuJoCo control tasks. These environments are widely adopted in continuous-control research due to their diverse dynamics, varying levels of difficulty, and well-established use in prior RL benchmarks.
We provide a comprehensive overview of these tasks and illustrate their corresponding state and action spaces in Fig.~\ref{task_visual}.

\textbf{HalfCheetah.}  
The halfcheetah robot embodies a two-dimensional structure designed to mimic the gait of a running cheetah. In the 17-dimensional state space, the robot encapsulates information such as positions, velocities, and motor forces across its body. The robot is capable of controlling its movements across six distinct action dimensions, thereby enabling efficient forward locomotion.

\textbf{Hopper.}
The hopper robot embodies a simplified one-legged structure designed for vertical locomotion. In the 11-dimensional state space, the robot encapsulates information encompassing joint angles and velocities, as well as the contact state with the ground. The robot orchestrates its balance and propulsion across 3 distinct action dimensions to effectively perform jumping movements.

\textbf{Walker2d.} 
The walker robot embodies a two-dimensional structure with 7 interconnected links representing two legs and a torso. Within the 17-dimensional state space, the robot encapsulates critical information such as joint positions and velocities. The robot manipulates its control across 6 distinct action dimensions to achieve articulated walking and stabilization.

\textbf{Ant.}
The ant robot embodies a complex four-legged structure featuring a body with multiple joints. In the extensive 111-dimensional state space, the robot encapsulates detailed positional, velocity, and force information across its entire frame. 
The robot has 8 joints for complex operations and is designed for flexible and stable quadrupedal locomotion.

\textbf{Humanoid.}
The humanoid robot embodies an intricate three-dimensional structure designed to mimic human bipedal locomotion, featuring a torso, arms, and legs. In the expansive 376-dimensional state space, the robot encapsulates extensive positional and velocity data across multiple body parts.
The robot achieves coordinated and complex bipedal balance and motion through 17 actuated joints, striving to attain human-like fluidity in its movements.

\textbf{Reacher.}  
The reacher robot embodies a simple robotic arm configuration designed for targeting tasks. In the 11-dimensional state space, the robot encapsulates joint angles and velocities, alongside the position of the target. The robot manipulates its reaching behavior across 2 distinct action dimensions, striving to minimize the distance between its end-effector and a designated target point.

\textbf{InvertedDoublePendulum.}
The inverted double pendulum robot embodies a dual-link structure designed for stabilization tasks under a pivot point. In the 11-dimensional state space, the robot encapsulates angular positions and velocities for both links. The robot modulates its stabilization control across 1 action dimension to maintain or regain an upright position.

\textbf{InvertedPendulum.}
The inverted pendulum robot embodies a classic single-link structure for studying balancing dynamics. In the compact 4-dimensional state space, the robot encapsulates the angular position and velocity of the pendulum, in tandem with the cart position and velocity. The robot directs its control across 1 action dimension, aiming to balance the pendulum atop the cart efficiently.

\textbf{Hyperparameters.}
To evaluate the IRA method against state-of-the-art online RL algorithms, we utilize the authors' implementations of ALH~\cite{quangaugmenting}, TD3~\cite{fujimoto2018addressing}, and PEER~\cite{he2023frustratingly}. Specifically, we employ the DDPG implementation from~\cite{fujimoto2018addressing} and the PyTorch-based PPO~\cite{schulman2017proximal} implementation. We build IRA on TD3~\cite{fujimoto2018addressing}.
A description of the MuJoCo tasks employed in the experiment can be found in the Table~\ref{tab:hyper}.

\begin{table}[!ht]
\caption{Hyperparameters for IRA.}
\label{tab:hyper}
\centering
\begin{tabular}{cll}
\toprule
                              & Hyperparameter          & \multicolumn{1}{l}{Value}           \\ \midrule
\multirow{13}{*}{IRA}         & Optimizer               & \multicolumn{1}{l}{Adam}            \\
                              & Critic learning rate    & \multicolumn{1}{l}{$3\times 10^{-4}$}            \\
                              & Actor learning rate     & \multicolumn{1}{l}{$3\times 10^{-4}$}  \\
                              & Batch size              & 256                                 \\
                              & Discount factor                & 0.99                                \\
                              & Number of iterations    & $1e6$                             \\
                              & Target update rate $\tau$     & 0.005                               \\
                              & Number of Critics & 2                                   \\
                               & RDE coefficient $\alpha$  & 5e-4  \\
                               & Action buffer size $n$ & 2e5  \\
                              & Number of neighbours $k$ & 10  \\
                              & Policy constraint coefficient $\mu$ & 1.0 $ \to $ 0.1  \\
                              & Policy update frequency $d$ & 1    \\
& Policy noise $\epsilon$   & 0.2  \\
                              & Noise clip range $c$ & [-0.5,0.5]    \\
                              \midrule
\multirow{2}{*}{Architecture} & Actor    & input-256-256-output                                 \\
                              & Critic & input-256-256-1                                      \\ \bottomrule
\end{tabular}
 
\end{table}
In Section~\ref{sec:method}, we introduce three parameters: an RDE coefficient $\alpha$=$5 \times 10^{-4}$, the number of action neighbors $k$=10, and a policy constraint strength $\mu$=1.0.
To ensure a fair comparison with the vanilla TD3, we adopt the noise parameters $\epsilon$ and $c$, as well as the hyperparameters $\tau$, directly from the TD3 author's implementation.
Unless otherwise stated, we maintain consistent hyperparameter configurations across all experiments. 
Regarding model architecture, our method faithfully replicates the structural framework of the TD3~\cite{fujimoto2018addressing} algorithm.
All experiments are conducted in a Linux environment equipped with a 56-core CPU and 2 Nvidia RTX 3090 Ti GPUs.

\textbf{Random Seeds.}
To ensure the reproducibility of IRA, we evaluated each algorithm using ten random seeds. Furthermore, we maintained consistent seeds across all experiments, applying them to PyTorch, Numpy, Gym, and CUDA packages.

\textbf{Evaluation.}
No data or parameters were repeated for training in the evaluation, and each evaluation step consisted of 10 randomly initialized rounds. We train for 1 million time steps and evaluate the policy every 5000 time steps. 
We present the mean and standard deviation of the final 10 trials.
\begin{figure*}[!ht]

\centering
\includegraphics[width=\textwidth]{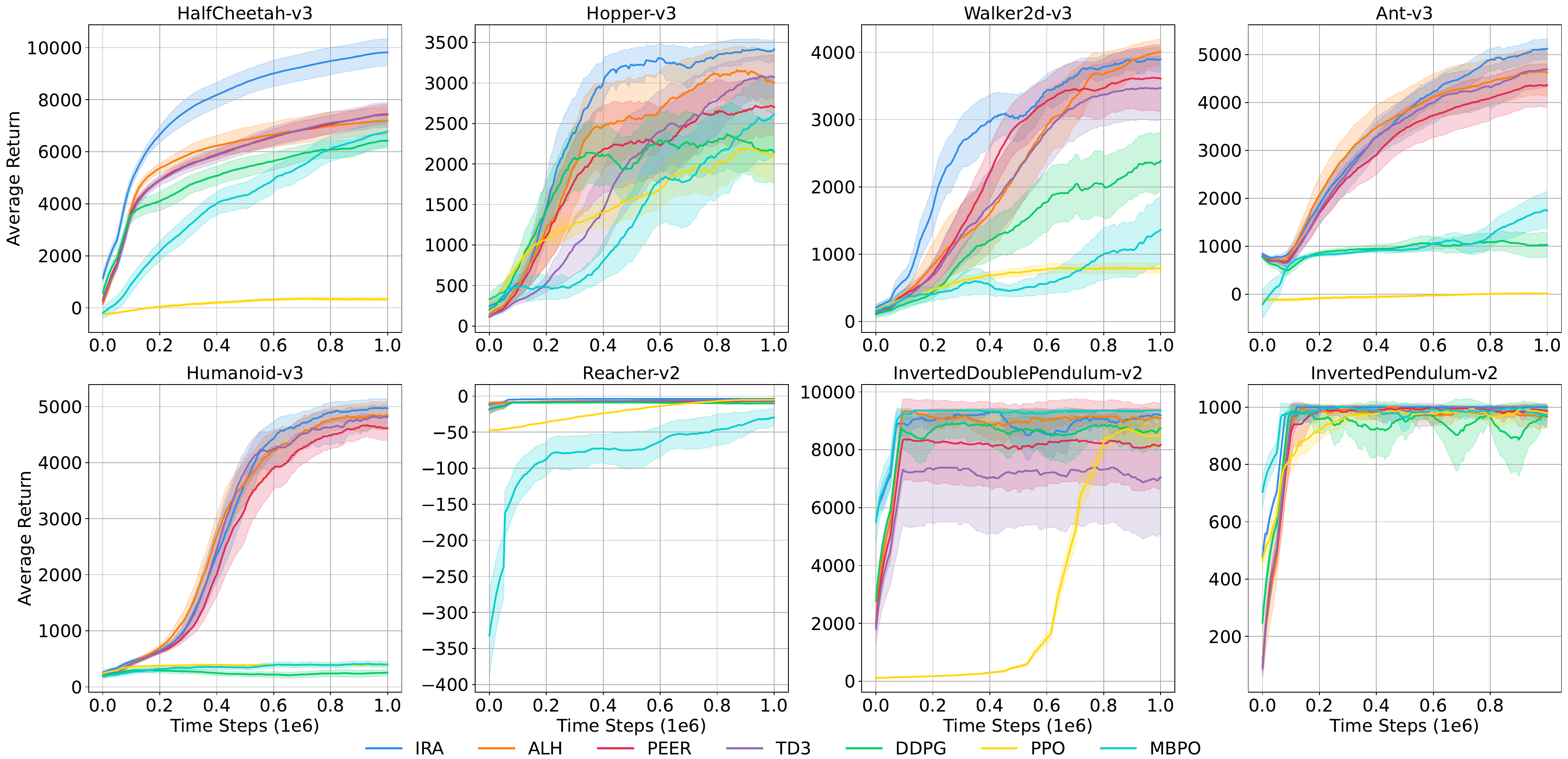}
\caption{Learning curves on four MuJoCo continuous control tasks. The shaded region represents half a standard deviation of the average evaluation over 10 trials. Curves are smoothed uniformly for visual clarity.}
\label{fig:Over_results}
 
\end{figure*}
\begin{table*}[!ht]
\centering
\setlength\tabcolsep{0.8pt}
\caption{Average return of the last 10 evaluation scores over 10 random seeds. The maximum values of each row are bolded. $\pm$ corresponds to a standard deviation over trials. The normalized score for each policy on a task is 100-scaled min-max normalization across the listed algorithms. ``InvertedDouble-v2" denotes the InvertedDoublePendulum-v2 task for brevity.}
\resizebox{1.\textwidth}{!}{
\begin{tabular}{lccccccc}
\toprule
Tasks        &IRA&ALH&PEER&TD3&DDPG&PPO&MBPO    \\  \midrule
HalfCheetah-v3&\bf 9832$\pm$517  &7202$\pm$527&7456$\pm$375&7442$\pm$477&6438$\pm$282& 334$\pm$68 &6782$\pm$554\\
Hopper-v3&\bf 3412$\pm$117&2993$\pm$402&2722$\pm$368&3079$\pm$260&2114$\pm$389& 2068$\pm$333&2671$\pm$476 \\
Walker2d-v3& 3886$\pm$193&\bf 4013$\pm$177 &3605$\pm$494&3464$\pm$478 &2379$\pm$420&791$\pm$67&1389$\pm$491\\
Ant-v3&\bf 5115$\pm$213 &4616$\pm$505&4360$\pm$450&4692$\pm$377&1031$\pm$262 &13$\pm$17&1739$\pm$382\\
Humanoid-v3&\bf 4963$\pm$166 &4742$\pm$229&4613$\pm$205&4843$\pm$203&258$\pm$54&394$\pm$15&406$\pm$60 \\
Reacher-v2&\bf -4$\pm$0  &-6$\pm$0&-7$\pm$0&-6$\pm$0&-10$\pm$1&-5$\pm$0&-30$\pm$12\\
InvertedDouble-v2& 9203$\pm$178& 8320$\pm$1408&8319$\pm$1407&7082$\pm$1918&8749$\pm$509&8506$\pm$397&\bf 9359$\pm$1\\
InvertedPendulum-v2&\bf 1000$\pm$0&980$\pm$24&983$\pm$25&987$\pm$20 &982$\pm$27&977$\pm$23&965$\pm$53  \\
 \midrule
Avg (normalized)& 98.7 &81.3&72.8&72.1&41.9&24.4&33.5\\
\bottomrule
\end{tabular}
}
\label{tab:mojoco_res}
\end{table*}
\subsection{Results of Empirical Evaluation}
We compare our proposed algorithm to state-of-the-art value-based algorithms TD3~\cite{fujimoto2018addressing} and DDPG~\cite{lillicrap2015continuous}, policy-based algorithm PPO~\cite{schulman2017proximal}, the representation learning-based algorithm ALH~\cite{quangaugmenting}, PEER~\cite{he2023frustratingly} and model-based algorithm (MBPO)~\cite{janner2019trust}.
All comparisons are conducted on eight continuous control tasks from the MuJoCo benchmark.
In Fig.~\ref{fig:Over_results}, our proposed method consistently outperforms the vanilla TD3 algorithm across different tasks, demonstrating remarkable advancement and stability compared to the baseline algorithm.

Specifically, our method achieves optimal performance compared to other algorithms in benchmark tasks such as HalfCheetah, Hopper, Ant, Humanoid, Reacher, and InvertedPendulum. In addition, our method also shows strong competitiveness in the Walker2d and InvertedDoublePendulum tasks, indicating that integrating our solution with a value-based algorithm does not degrade the overall performance.
As shown in Fig.~\ref{fig:Over_results}, value-based algorithms such as TD3 and DDPG converge more slowly than our proposed IRA, demonstrating the better policy exploitation ability of the IRA. Moreover, our proposed method outperforms representation learning-based algorithms (ALH and PEER), policy-based algorithm (PPO), and model-based algorithm (MBPO).
As shown in Table~\ref{tab:mojoco_res}, specifically, compared with the benchmark algorithms including ALH, PEER, TD3, DDPG, PPO, and MBPO, the IRA exhibits superior performance in the average evaluation metrics, achieving relative improvements of 21.4\%, 35.6\%, 36.9\%, 135.6\%, 304.5\%, and 194.6\%, respectively.
Moreover, the proposed IRA method demonstrates improved stability over the vanilla TD3 algorithm, as evidenced by lower performance variance. This indicates that using the nearest-neighbor policy as an update anchor in IRA effectively contributes to more stable performance improvement.

\begin{figure*}[!ht]

\centering
\includegraphics[width=\textwidth]{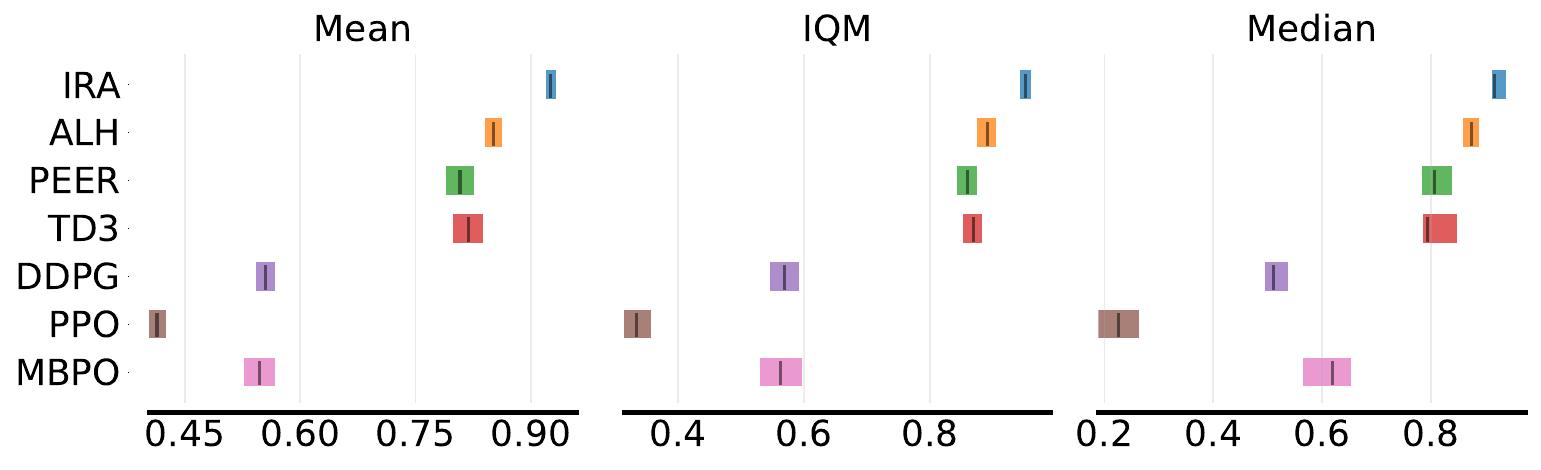}
\caption{Aggregate performance of IRA across eight continuous control tasks using the RLiable analysis framework, reporting Mean, IQM, and Median metrics over the entire empirical evaluation.
}
\label{fig:iqm_mean}
 
\end{figure*}

Additionally, we performed a meta-analysis of our empirical evaluation using the RLiable~\cite{agarwal2021deep} analysis framework, as shown in Fig.~\ref{fig:iqm_mean}.
IRA consistently outperforms baseline methods across all three RLiable metrics. It achieves the highest Mean score, indicating superior overall returns, and leads in both IQM and Median, reflecting its robustness and stability across diverse continuous control tasks. These results demonstrate that IRA not only attains strong peak performance but also maintains consistent and reliable performance. This level of performance underscores IRA's practical advantage over existing RL algorithms and suggests its strong potential for advancing the deployment of RL in real-world engineering applications.

\subsection{Ablation Study}
\textbf{Effectiveness of Q-Representation Discrepancy Evolution.}
To analyze the effect of different $\alpha$ values of the RDE on the IRA architecture, we constructed a set of experiments for this analysis. We evaluated them on the HalfCheetah and Ant tasks using $\alpha$ values of $(5e-5, 5e-4, 5e-3, 5e-2)$ respectively. Subsequently, we compared these results with those obtained by an IRA without the RDE component (IRA(w/o RDE)). We computed the means and standard deviations of the last ten evaluation scores, and the results are presented in Table~\ref{ab_rde}. 
Experimental results demonstrate that when $\alpha$=5e-3, $\alpha$=5e-4, or $\alpha$=5e-5, the Q-representation discrepancy evolution approach achieves higher average scores compared to the IRA(w/o RDE). 
This highlights the role of RDE in improving the discriminability of adjacent action representations, thereby further enhancing the performance of greedy action guidance. 
However, the performance of the HalfCheetah and Ant tasks drops significantly when $\alpha$=5e-2. We consider that increased representation constraints can induce fluctuations within the Q-network, resulting in unstable estimates of Q-values.

Furthermore, Table~\ref{ab_rde} shows that IRA(w/o RDE) consistently results in performance degradation compared to IRA in HalfCheetah and Ant tasks, further attesting to the effectiveness of our approach. 
In particular, we observe a more significant performance drop in IRA(w/o RDE) compared to the IRA in the Ant task. 
We attribute this to the higher complexity of the Ant task, where the state space has 111 dimensions and the action space has 8 dimensions. This increases the necessity of learning representations that can effectively distinguish between neighbor actions.
\begin{table}[!htpb]
\centering
\setlength\tabcolsep{1.9pt}
\caption{Sensitivity experiment for parameter $\alpha$.}
\begin{tabular}{lcc}
\toprule
Method         &HalfCheetah-v3 &Ant-v3   \\ 
\midrule
IRA(w/o RDE)&  9453$\pm$520&3643$\pm$435 \\
IRA($\alpha=5e-5$)&10011$\pm$696  &4975$\pm$335  \\
IRA($\alpha=5e-4$)&   9832$\pm$517&5115$\pm$213 \\
IRA($\alpha=5e-3$)&10384$\pm$421  &4672$\pm$371  \\
IRA($\alpha=5e-2$)&9134$\pm$760 & 3957$\pm$407  \\
\bottomrule
\end{tabular}
\label{ab_rde}
\end{table}

\begin{table}[h]
\centering
\caption{Performance comparison between the RDE method and contrastive learning approaches on IRA, averaged over 10 random seeds.}
\begin{tabular}{lll}
\toprule
Method& HalfCheetah-v3  &Hopper-v3     \\
\midrule
IRA(RDE)& 9832$\pm$517 &3412$\pm$117\\
IRA(CL)& 9779$\pm$643 &3325$\pm$283\\
 \bottomrule
\end{tabular}
\label{tab:CL}
\end{table}
We compare the performance of the RDE method with the IRA method combined with contrastive learning (CL)~\cite{oord2018representation}.
The action with the highest value is selected from the $k$ nearest neighbor actions as the positive sample, while suboptimal actions are treated as negative samples to construct the contrastive learning loss.
In the experiments, the regularization loss weight is set to $\alpha$=5e-4, as shown in Table~\ref{tab:CL}. CL reduces the representation discrepancy between the current policy and the highest-value action (positive sample); however, the current policy itself may also be suboptimal, which can introduce training instability and lead to performance degradation in the Hopper task.

\begin{figure*}[!ht]
\centering
\includegraphics[width=\textwidth]{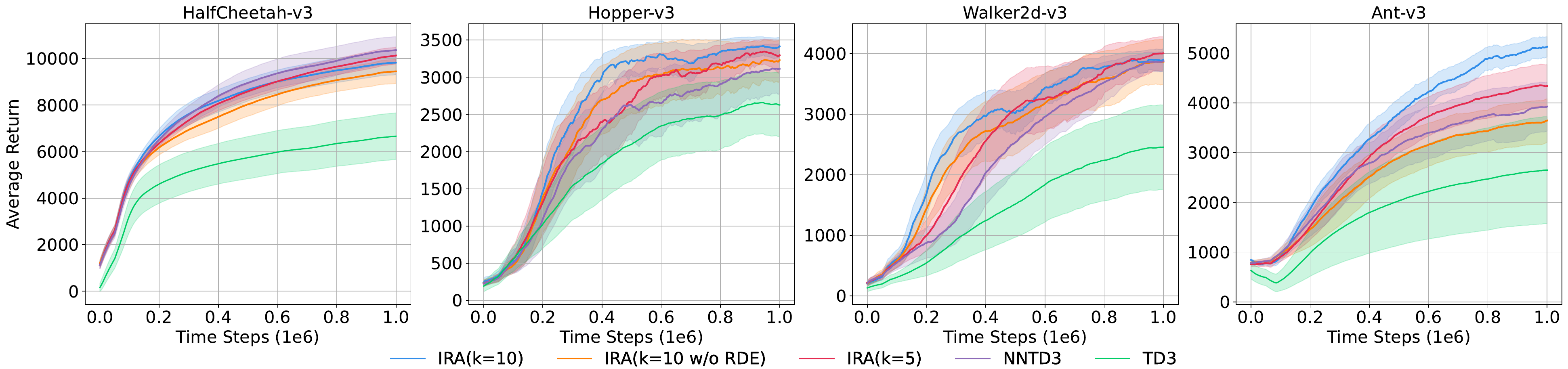}
\caption{Learning curves obtained with varying $k$ across four MuJoCo continuous control tasks. The shaded region represents half a standard deviation of the average evaluation over 10 trials. Curves are smoothed uniformly for visual clarity.}
\label{ab_study_k}
 
\end{figure*}
\textbf{Effectiveness of Greedy Action Guidance.}
We conducted a systematic ablation analysis on MuJoCo environments to evaluate four configuration variants, aiming to quantify the contribution of greedy action guidance in the IRA algorithm.
In the first configuration, we demonstrate greedy action guidance by assessing the optimal action within the neighborhood when setting the IRA parameter to ($k$=10).
The second configuration employs the IRA algorithm without the Q-representation discrepancy evolution approach, which is denoted as IRA ($k$=10 w/o RDE). 
In the third configuration, we reduce the value of $k$ to 5 (IRA $k$=5) to evaluate its impact on the IRA method.
Finally, we implement the nearest-neighbor exploration algorithm proposed in NNAC~\cite{shen2021theoretically} based on the TD3 algorithm, referred to as NNTD3. NNTD3 employs the nearest neighbor action from the online replay buffer as an anchor for the current policy decision and adopts a supervised learning paradigm to guide the actor update. The overall actor loss can be formulated as
\begin{equation}
    \label{nn_exploreation_baseline1}
    \begin{aligned}
J_{\pi}(\phi)= {\mathbb{E}_{s}}\left[ {{Q_\theta }(s,{\pi _\phi }(s)) - \mu {{({\pi _\phi }(s) - {{\tilde a}_{opt}})}^2}} \right],\\
{\tilde a}_{opt} =  \argmin_{a} \left(\max(|\pi (s)_j - a_j|)\right),
\end{aligned}
\end{equation}
where $a \in {\mathcal{A}}$ denotes iterating over all actions in the explored action buffer.
In these setups, we both rely on Q-values and optimal action neighbor to guide the policy update process.

The results in Fig.~\ref{ab_study_k} demonstrate that IRA consistently enhances policy performance. We observe that the variant without RDE (IRA, $k{=}10$ w/o RDE) shows a clear degradation compared to the full IRA method. This gap highlights that enlarging the representation discrepancy between optimal and suboptimal actions enables the Q-network to more reliably identify high-value policy anchors, ultimately leading to stronger performance gains.
As the value of $k$ increases, the greater exploration of deterministic guidance results in improved performance, a trend consistently observed in the Hopper, Walker2d, and Ant tasks.
Taken together, these findings suggest that choosing $k \in \{5,10\}$ provides a stable and effective range for practical deployment.
In these tasks, we observe that nearest action exploitation (NNTD3) significantly enhances the performance of the vanilla TD3 algorithm. This suggests that leveraging neighboring actions as an update anchor for policy updates effectively addresses the uncertainties arising from Q-guidance.
However, in the HalfCheetah task, we observed that nearest action exploitation (NNTD3) achieved superior performance.
This can be attributed to the fact that larger $k$ values involve evaluating a greater number of actions, which propagate distantly spaced anchors and result in potentially unstable learning dynamics.

\begin{figure*}
\centering
\includegraphics[width=\textwidth]{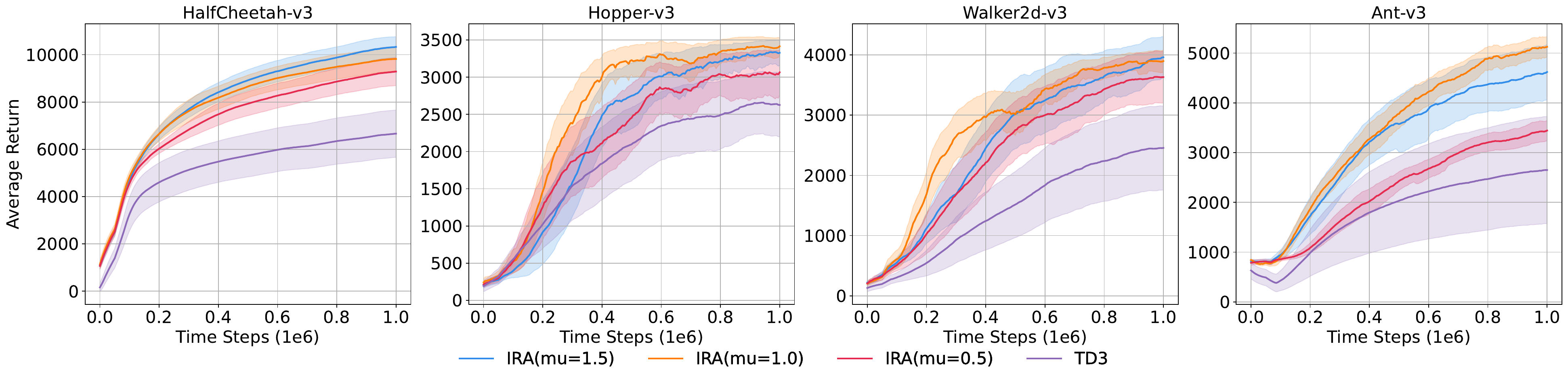}
\caption{Learning curves obtained with varying $\mu$ across four MuJoCo continuous control tasks. The shaded region represents half a standard deviation of the average evaluation over 10 trials. Curves are smoothed uniformly for visual clarity.}
\label{ablation_mu_all}
\end{figure*}

To further evaluate the sensitivity of the IRA algorithm to the hyperparameter $\mu$, we conduct experiments during the training phase with three different values of $\mu=(0.5, 1.0, 1.5)$. Each model is trained for 1 million time steps on the first four MuJoCo tasks, using 10 random seeds. The results are presented in Fig.~\ref{ablation_mu_all}.  
The findings demonstrate that our proposed method significantly outperforms the baseline in terms of performance and exhibits superior sample efficiency. Notably, the performance decreases when $\mu$=0.5, suggesting that a weaker level of policy constraint is insufficient to guide policy optimization effectively. Interestingly, when $\mu$=1.5, the IRA achieves better performance on the HalfCheetah and Walker2d tasks, while $\mu$=1.0 yields better performance on the Hopper and Ant tasks. These outcomes indicate that the optimal choice for the policy constraint strength $\mu$ may vary depending on the nature of the specific task. By tuning $\mu$ to better suit the characteristics of each task, higher performance can be achieved.  

\begin{figure}
\centering
\includegraphics[width=0.4\textwidth]{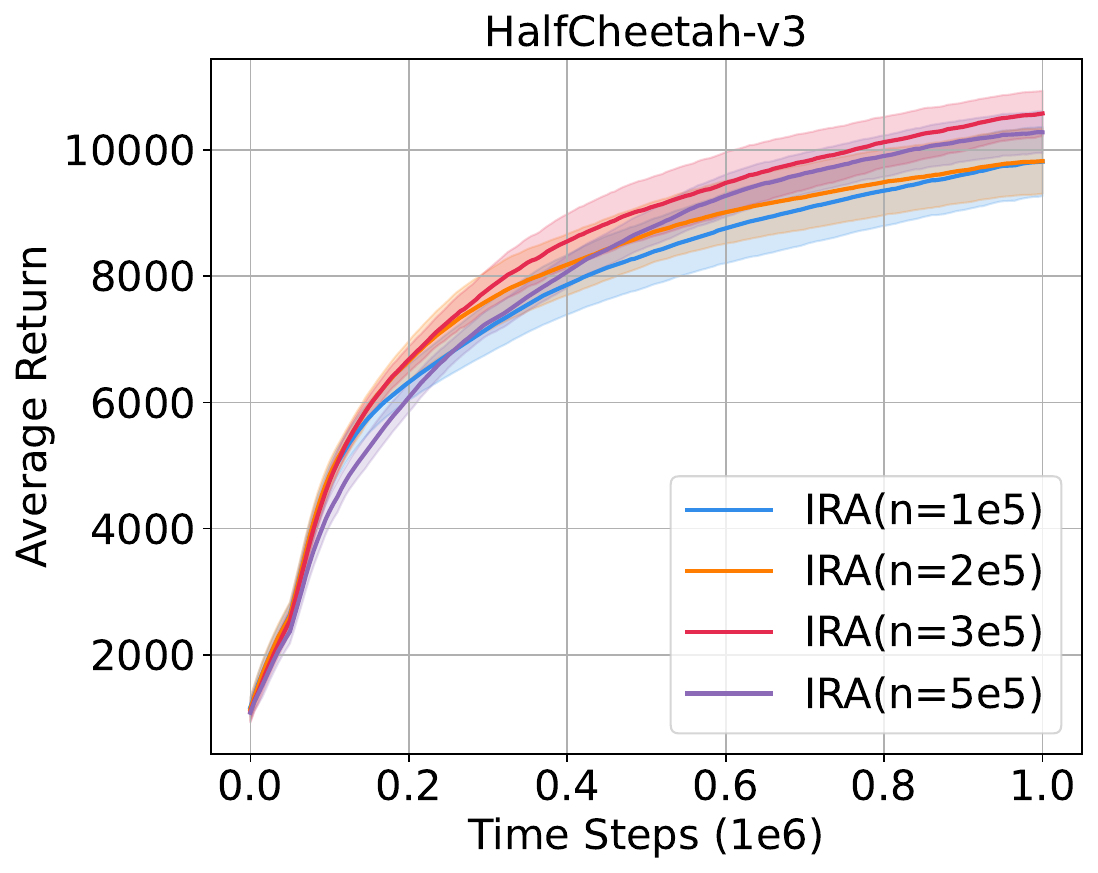}
\caption{Learning curves obtained with varying action buffer size $n$ across four MuJoCo continuous control tasks. The shaded region represents half a standard deviation of the average evaluation over 10 trials. Curves are smoothed uniformly for visual clarity.}
\label{ablation_n}
\end{figure}


We further conducted experiments to examine the impact of action buffer size \( n \) on the HalfCheetah task, using \( n \) values of (1e5, 2e5, 3e5, 5e5). The experimental results are presented in Fig.~\ref{ablation_n}. It can be observed that as the action buffer size \( n \) increases, the performance improves significantly, which exhibits a notable performance enhancement. This indicates that the IRA method can achieve greater performance gains in RL training by retrieving historical actions.
However, we observe that when $n$=5e5, the performance is lower than when $n$=3e5. This indicates that storing an excessive number of actions introduces a large proportion of low-quality actions; therefore, selecting an appropriately sized action buffer or evaluating more nearest neighbor actions is crucial.

\begin{figure*}[!ht]
\centering
 \includegraphics[width=0.75\textwidth]{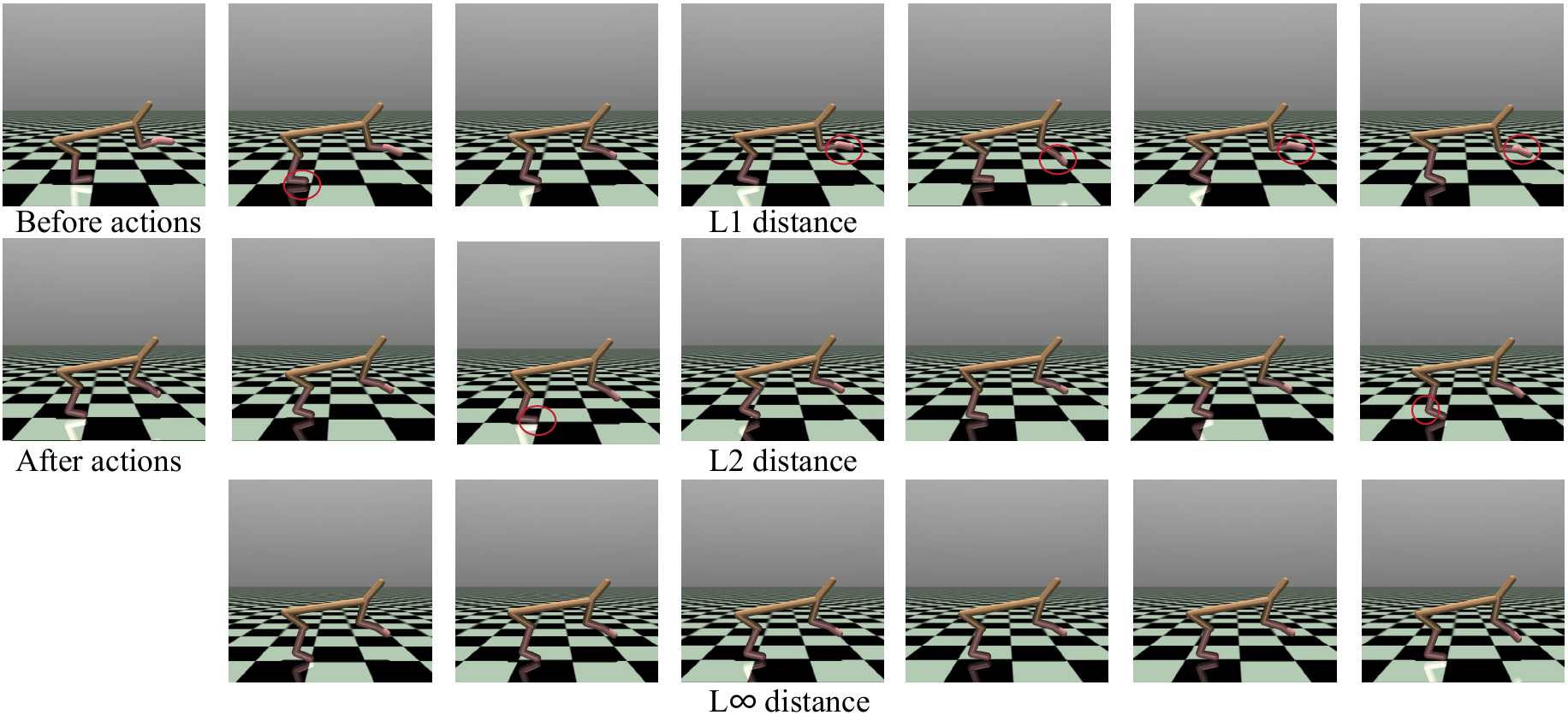} 
 \caption{Comparison of discrepancies between retrieved actions under different distance metrics and the actual states induced by policy execution. The $L_1$ and $L_2$ norms are more sensitive to subtle variations in robot joint movements, whereas the $L_\infty$ norm retrieves nearest neighbor actions that more closely match the current decision state. Joint dimensions exhibiting large deviations from the ``After actions'' are highlighted with red circles.}
  \label{metrix_l1_l2_l_infy}
\end{figure*}
Additionally, we visualize the nearest neighbor actions retrieved using different distance metrics ($L_1$, $L_2$, $L_\infty$) and execute the resulting states on the ``Before actions'', with the results presented in Fig.~\ref{metrix_l1_l2_l_infy}. The actions retrieved using the $L\infty$ distance metric exhibit the highest similarity to the next state predicted by the current policy (After actions), indicating that focusing on larger changes ($L_\infty$) in the action dimensions is highly effective.

\begin{center}  
 
\begin{minipage}{.35\textwidth} 
\begin{algorithm}[H]
   \caption{Policy Update Mechanism}
   \label{alg:policy_update}
\begin{algorithmic}
   \STATE Initialize timestamp $t$=0, $d$=1 or 2
   \WHILE {True}
   \STATE Update timestamp $t$=$t$+1
   \STATE Train critics
   \IF {$t$\%\textit{d}==0} 
   \STATE Train actor
   \STATE Update weights
   \ENDIF
   \ENDWHILE
\end{algorithmic}
\label{ipu_mechin}
\end{algorithm}
\end{minipage}
\end{center}
 
\begin{figure*}
\includegraphics[width=\textwidth]{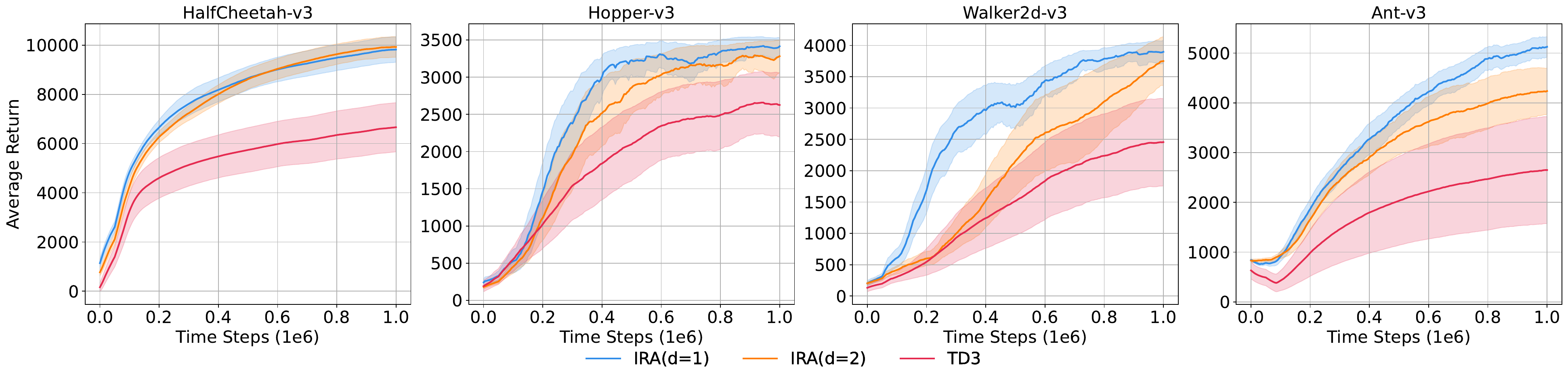}
\caption{Learning curves obtained with varying $d$ across four MuJoCo continuous control tasks. The shaded region represents half a standard deviation of the average evaluation over 10 trials. Curves are smoothed uniformly for visual clarity.}
\label{ablation_d}
\end{figure*}
\textbf{Effectiveness of the Instant Policy Updates.}
To validate the effectiveness of the instant policy update mechanism, we compare the IRA algorithm under different policy update settings, specifically instant policy update ($d$=1) and delayed policy update ($d$=2) during training, as detailed in Algorithm~\ref{ipu_mechin}.
The results are illustrated in Fig.~\ref{ablation_d}, our proposed method significantly outperforms the vanilla TD3 algorithm, achieving faster convergence compared to $d$=2. Notably, the IRA demonstrates superior performance on the Ant task when $d$=1. 
The inferior performance observed when $d$=2 stems from the low policy update frequency of the IRA algorithm, which leads to inefficient sample utilization. In contrast, when $d$=1, the accelerated policy update frequency effectively alleviates this limitation by enabling rapid policy improvement to adapt to dynamic and continuously changing environments.
In the HalfCheetah task, the performance of $d$=2 significantly outperforms that of $d$=1. We attribute this improvement to the stabilization of policy optimization in the later stages of training, which necessitates a reduction in the frequency of policy updates to promote policy stability.

 Importantly, when $d$=1, the IRA exhibits enhanced performance in the Hopper, Walker2d, and Ant tasks. We contend that the increased frequency of policy updates effectively mitigates the issue of slow policy exploitation, facilitating rapid policy enhancements that are better suited for dynamic and continually changing environments.

\begin{figure}[!ht]
\centering
\includegraphics[width=0.5\textwidth]{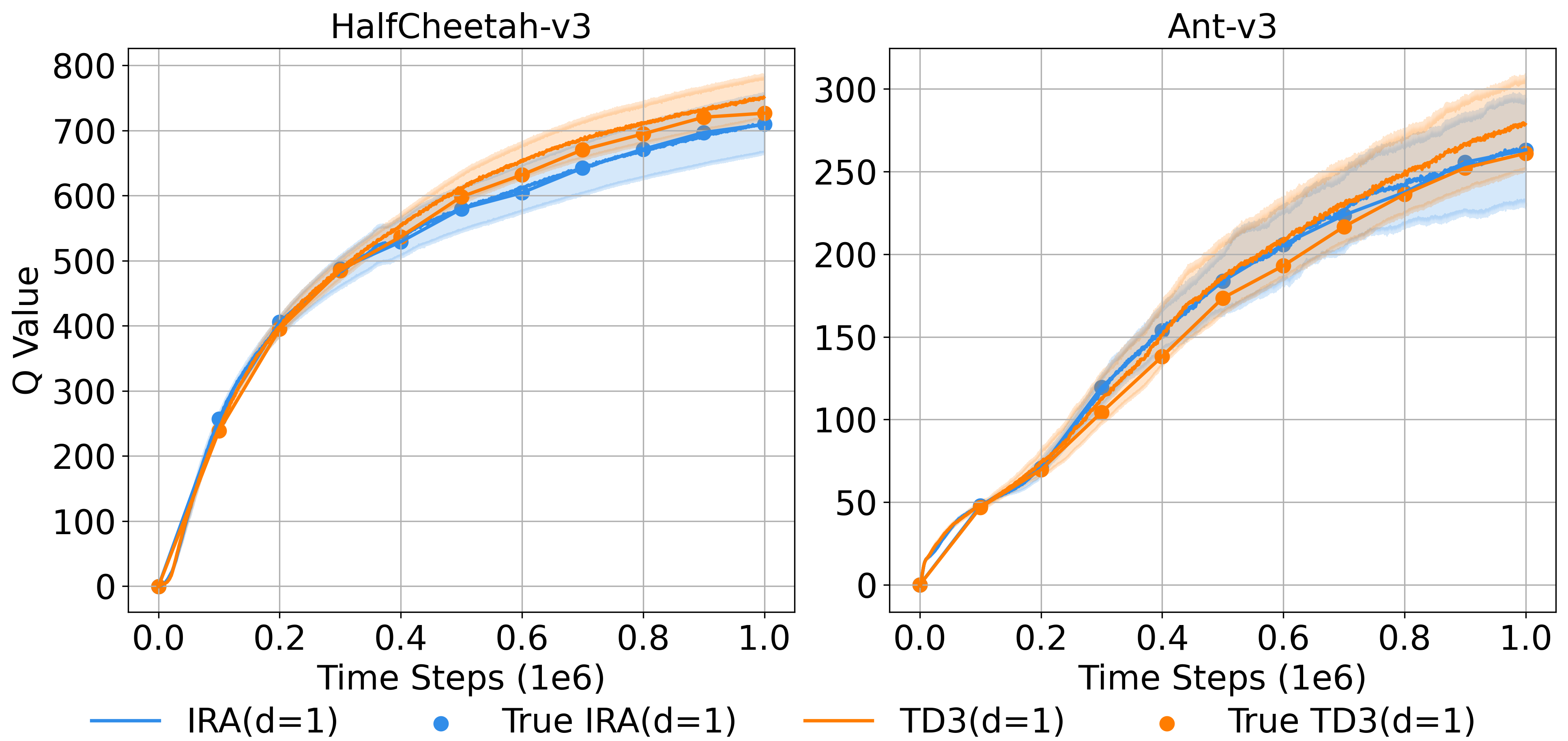}
\caption{The overestimation bias of value estimates in the IRA and TD3 algorithms with the IPU mechanism is measured over 1 million time steps.}
\label{ablation_overestimation}
 
\end{figure}
\subsection{Mitigating Value Overestimation with IRA}
To assess the robustness of IRA against overestimation bias, we experimentally compare the variations in predicted Q-values and true Q-values using the TD3 algorithm with $d=1$ and the IRA ($d$=1) algorithm, respectively, as shown in Fig.~\ref{ablation_overestimation}.
As illustrated in Fig.~\ref{ablation_overestimation}, the TD3 algorithm with instant policy updating ($d$=1) exhibits a significant estimation error, which progressively increases with each training step. In contrast, the IRA algorithm effectively mitigates this accumulation of error. Notably, the instant policy updates within the IRA lead to a distribution of predicted Q-values that closely approximates the true Q-values. We argue that the implementation of greedy action guidance employs constrained exploration, which partially alleviates the issue of Q-value overestimation.

In addition, the exploration of nearest optimal action functions as a robust policy constraint, thereby preventing the learning algorithm from venturing into regions with high uncertainty. This underscores the argument that greedy action guidance not only tackles the issue of slow exploitation, enhancing the overall performance of the model, but also represents a pivotal approach for alleviating Q-value overestimation concerns.

\begin{figure*}[!ht] 
\centering
\includegraphics[width=\textwidth]{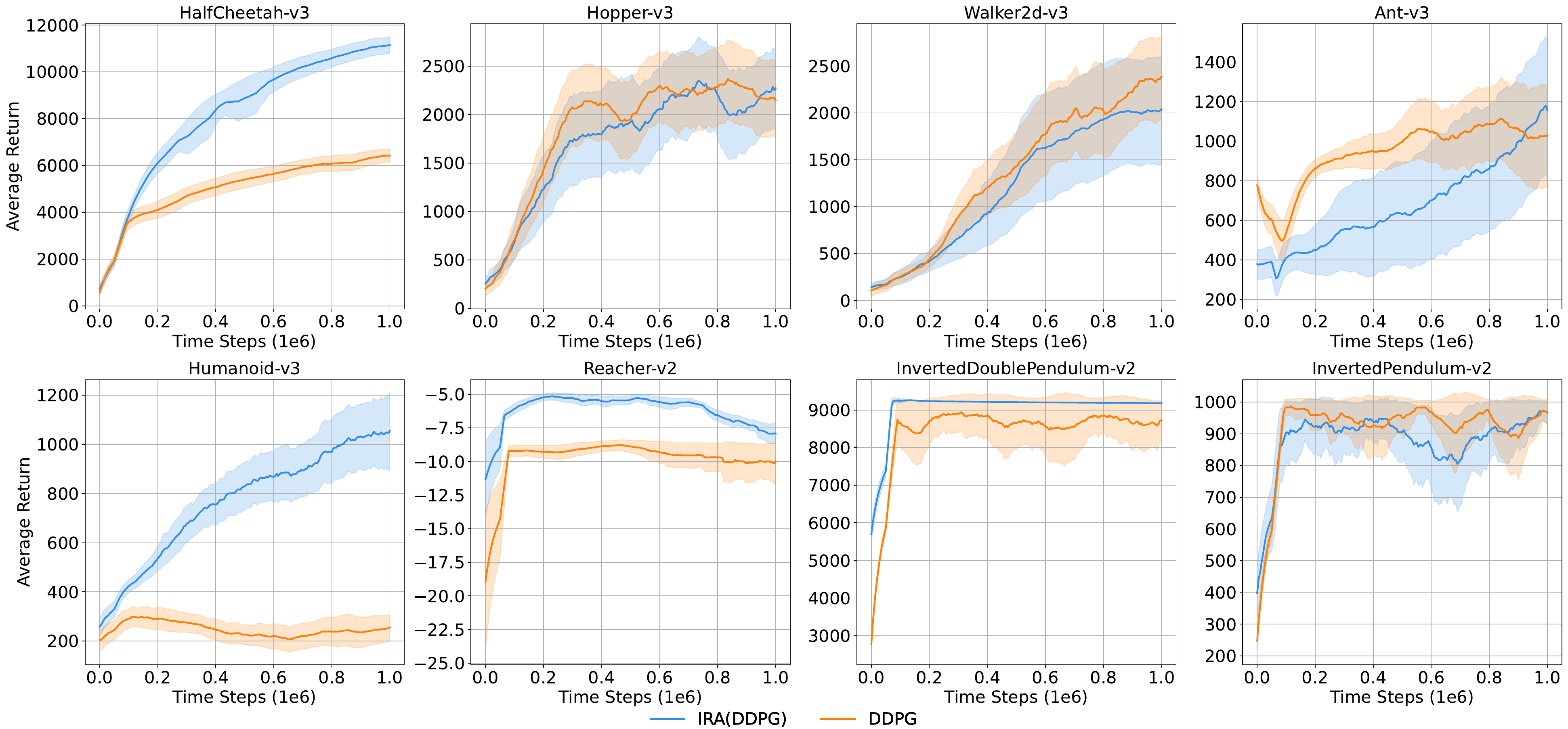}
\caption{Learning curves on four MuJoCo continuous control tasks. The shaded region represents half a standard deviation of the average evaluation over 10 trials. Curves are smoothed uniformly for visual clarity.}
\label{fig:Over_results_ddpg}
\end{figure*}
\subsection{Combined with Other Value-based Algorithm}
\label{sec:ddpg_ira}
Moreover, to further demonstrate the effectiveness of IRA in alleviating the slow exploitation issue in value-based RL, and to provide stronger evidence that this problem is not specific to TD3, we additionally implemented IRA on the DDPG algorithm and conducted corresponding experiments. Since Q-value is the only guide for the policy, vanilla DDPG also relies on vanilla Bellman optimization to update the policy. We conducted experiments in 8 MuJoCo environments with 10 trials each, all under the same hyperparameter settings and operating conditions. Fig.~\ref{fig:Over_results_ddpg} compares the learning curves of our IRA with those of the DDPG. 
 The DDPG version of IRA demonstrates notably superior performance compared to vanilla DDPG in several aspects: in four out of the eight environments, the IRA exhibits significant performance improvements, while maintaining comparable performance to vanilla DDPG in the remaining environments. Specifically, in the HalfCheetah, Humanoid, Reacher, and InvertedDoublePendulum environments, the IRA significantly outperforms the vanilla DDPG.

\begin{table*}[!ht]
\centering
\caption{Time consumption throughout the entire training process of the algorithm.}
\begin{tabular}{lccccccc}
\toprule
Methods        &IRA&IRA(DDPG)&ALH&PEER&TD3&DDPG&PPO    \\  
 \midrule
runtime&6.5h& 5.9h &3.4h&3.3h&2.7h&3.1h&4.1h\\
\bottomrule
\end{tabular}
\label{tab:consumer_time}
\end{table*}
\subsection{Runtime}
We conducted experiments in a Linux environment equipped with a 56-core CPU and 2 Nvidia RTX 3090 Ti GPUs to evaluate the computational time of our proposed IRA method against ALH, PEER, TD3, DDPG, and PPO. As illustrated in Table \ref{tab:consumer_time}, our IRA method significantly increased the training time compared to vanilla algorithms, with both IRA and IRA(DDPG) showing considerable time consumption. This additional time overhead is primarily attributed to the retrieval process within the action buffer.


\section{Conclusions and Future Work}
\label{conclusion}
In this work, we undertake a comprehensive investigation of representation learning algorithms and policy constraint algorithms designed to enhance policy exploitation. 
We propose the IRA algorithm, which introduces additional representation supervision signals to learn distinguishable action representations and leverages optimal nearest-neighbor actions as anchors for policy updates, thereby improving sample efficiency. Moreover, IRA incorporates an instant policy update mechanism to further enhance policy exploitation.
We implement the IRA algorithm based on both TD3 and DDPG frameworks. Extensive empirical evaluations under fixed parameter settings demonstrate that IRA consistently outperforms the corresponding baseline algorithms across a variety of continuous control tasks, and further surpasses other representation learning and policy constraint methods.
Ablation studies indicate that the instant retrospect action mechanism introduced in the learning process of IRA effectively enhances policy exploitation and mitigates the overestimation bias of the Q-function.
This work aims to improve the sample efficiency of value-based RL. Extending IRA to the maximum entropy RL framework remains a promising direction for future research.

\bibliographystyle{IEEEtran}
 
\bibliography{ira}

@inproceedings{fujimoto2018addressing,
  title={Addressing function approximation error in actor-critic methods},
  author={Fujimoto, Scott and Hoof, Herke and Meger, David},
  booktitle={International Conference on Machine Learning},
  pages={1587--1596},
  year={2018},
  organization={PMLR}
}

@article{lillicrap2015continuous,
  title={Continuous control with deep reinforcement learning},
  author={Lillicrap, TP},
  journal={arXiv preprint arXiv:1509.02971},
  year={2015}
}

@article{schulman2017proximal,
  title={Proximal policy optimization algorithms},
  author={Schulman, John and Wolski, Filip and Dhariwal, Prafulla and Radford, Alec and Klimov, Oleg},
  journal={arXiv preprint arXiv:1707.06347},
  year={2017}
}

@article{janner2019trust,
  title={When to trust your model: Model-based policy optimization},
  author={Janner, Michael and Fu, Justin and Zhang, Marvin and Levine, Sergey},
  journal={Advances in Neural Information Processing Systems},
  volume={32},
  year={2019}
}

@article{agarwal2021deep,
  title={Deep reinforcement learning at the edge of the statistical precipice},
  author={Agarwal, Rishabh and Schwarzer, Max and Castro, Pablo Samuel and Courville, Aaron C and Bellemare, Marc},
  journal={Advances in Neural Information Processing Systems},
  volume={34},
  pages={29304--29320},
  year={2021}
}

@inproceedings{he2023frustratingly,
  title={Frustratingly easy regularization on representation can boost deep reinforcement learning},
  author={He, Qiang and Su, Huangyuan and Zhang, Jieyu and Hou, Xinwen},
  booktitle={Proceedings of the IEEE/CVF conference on computer vision and pattern recognition},
  pages={20215--20225},
  year={2023}
}

@inproceedings{quangaugmenting,
  title={Augmenting decision with hypothesis in reinforcement learning},
  author={Quang, Nguyen Minh and Lauw, Hady W},
  booktitle={International Conference on Machine Learning},
  year={2024}
}

@article{shah2018q,
  title={Q-learning with nearest neighbors},
  author={Shah, Devavrat and Xie, Qiaomin},
  journal={Advances in Neural Information Processing Systems},
  volume={31},
  year={2018}
}

@inproceedings{cheng2022adversarially,
  title={Adversarially trained actor critic for offline reinforcement learning},
  author={Cheng, Ching-An and Xie, Tengyang and Jiang, Nan and Agarwal, Alekh},
  booktitle={International Conference on Machine Learning},
  pages={3852--3878},
  year={2022},
  organization={PMLR}
}

@inproceedings{fujimoto2021minimalist,
  title={A minimalist approach to offline reinforcement learning},
  author={Fujimoto, Scott and Gu, Shixiang Shane},
  booktitle={Advances in neural information processing systems},
  volume={34},
  pages={20132--20145},
  year={2021}
}

@article{kumar2023policy,
  title={Policy gradient for rectangular robust markov decision processes},
  author={Kumar, Navdeep and Derman, Esther and Geist, Matthieu and Levy, Kfir Y and Mannor, Shie},
  journal={Advances in Neural Information Processing Systems},
  volume={36},
  pages={59477--59501},
  year={2023}
}

@article{ma2023iteratively,
  title={Iteratively refined behavior regularization for offline reinforcement learning},
  author={Ma, Yi and Hao, Jianye and Hu, Xiaohan and Zheng, Yan and Xiao, Chenjun},
  journal={Advances in Neural Information Processing Systems},
  volume={37},
  pages={56215--56243},
  year={2024}
}

@article{tarasov2024revisiting,
  title={Revisiting the minimalist approach to offline reinforcement learning},
  author={Tarasov, Denis and Kurenkov, Vladislav and Nikulin, Alexander and Kolesnikov, Sergey},
  journal={Advances in Neural Information Processing Systems},
  volume={36},
  year={2024}
}

@inproceedings{ran2023policy,
  title={Policy regularization with dataset constraint for offline reinforcement learning},
  author={Ran, Yuhang and Li, Yi-Chen and Zhang, Fuxiang and Zhang, Zongzhang and Yu, Yang},
  booktitle={International Conference on Machine Learning},
  pages={28701--28717},
  year={2023},
  organization={PMLR}
}

@book{deza2009encyclopedia,
  title={Encyclopedia of distances},
  author={Deza, Elena and Deza, Michel Marie and Deza, Michel Marie and Deza, Elena},
  year={2009},
  publisher={Springer}
}

@inproceedings{yarats2021improving,
  title={Improving sample efficiency in model-free reinforcement learning from images},
  author={Yarats, Denis and Zhang, Amy and Kostrikov, Ilya and Amos, Brandon and Pineau, Joelle and Fergus, Rob},
  booktitle={Proceedings of the AAAI Conference on Artificial Intelligence},
  volume={35},
  pages={10674--10681},
  year={2021}
}

@article{ni2023transformers,
  title={When do transformers shine in rl? decoupling memory from credit assignment},
  author={Ni, Tianwei and Ma, Michel and Eysenbach, Benjamin and Bacon, Pierre-Luc},
  journal={Advances in Neural Information Processing Systems},
  volume={36},
  pages={50429--50452},
  year={2023}
}

@article{fujimoto2024sale,
  title={For sale: State-action representation learning for deep reinforcement learning},
  author={Fujimoto, Scott and Chang, Wei-Di and Smith, Edward and Gu, Shixiang Shane and Precup, Doina and Meger, David},
  journal={Advances in Neural Information Processing Systems},
  volume={36},
  year={2024}
}

@article{he2022reinforcement,
  title={Reinforcement learning with automated auxiliary loss search},
  author={He, Tairan and Zhang, Yuge and Ren, Kan and Liu, Minghuan and Wang, Che and Zhang, Weinan and Yang, Yuqing and Li, Dongsheng},
  journal={Advances in Neural Information Processing Systems},
  volume={35},
  pages={1820--1834},
  year={2022}
}

@article{ni2024bridging,
  title={Bridging State and History Representations: Understanding Self-Predictive RL},
  author={Ni, Tianwei and Eysenbach, Benjamin and Seyedsalehi, Erfan and Ma, Michel and Gehring, Clement and Mahajan, Aditya and Bacon, Pierre-Luc},
  journal={arXiv preprint arXiv:2401.08898},
  year={2024}
}

@article{allen2021learning,
  title={Learning markov state abstractions for deep reinforcement learning},
  author={Allen, Cameron and Parikh, Neev and Gottesman, Omer and Konidaris, George},
  journal={Advances in Neural Information Processing Systems},
  volume={34},
  pages={8229--8241},
  year={2021}
}

@inproceedings{zheng2023adaptive,
  title={Adaptive policy learning for offline-to-online reinforcement learning},
  author={Zheng, Han and Luo, Xufang and Wei, Pengfei and Song, Xuan and Li, Dongsheng and Jiang, Jing},
  booktitle={Proceedings of the AAAI Conference on Artificial Intelligence},
  volume={37},
  number={9},
  pages={11372--11380},
  year={2023}
}

@inproceedings{van2016deep,
  title={Deep reinforcement learning with double q-learning},
  author={Van Hasselt, Hado and Guez, Arthur and Silver, David},
  booktitle={Proceedings of the AAAI Conference on Artificial Intelligence},
  volume={30},
  number={1},
  year={2016}
}

@article{mnih2013playing,
  title={Playing atari with deep reinforcement learning},
  author={Mnih, Volodymyr},
  journal={arXiv preprint arXiv:1312.5602},
  year={2013}
}

@inproceedings{espeholt2018impala,
  title={Impala: Scalable distributed deep-rl with importance weighted actor-learner architectures},
  author={Espeholt, Lasse and Soyer, Hubert and Munos, Remi and Simonyan, Karen and Mnih, Vlad and Ward, Tom and Doron, Yotam and Firoiu, Vlad and Harley, Tim and Dunning, Iain and others},
  booktitle={International Conference on Machine Learning},
  pages={1407--1416},
  year={2018},
  organization={PMLR}
}

@article{mnih2016asynchronous,
  title={Asynchronous Methods for Deep Reinforcement Learning},
  author={Mnih, Volodymyr},
  journal={arXiv preprint arXiv:1602.01783},
  year={2016}
}

@inproceedings{shen2021theoretically,
  title={Theoretically principled deep RL acceleration via nearest neighbor function approximation},
  author={Shen, Junhong and Yang, Lin F},
  booktitle={Proceedings of the AAAI Conference on Artificial Intelligence},
  volume={35},
  number={11},
  pages={9558--9566},
  year={2021}
}

@inproceedings{yu2023actor,
  title={Actor-critic alignment for offline-to-online reinforcement learning},
  author={Yu, Zishun and Zhang, Xinhua},
  booktitle={International Conference on Machine Learning},
  pages={40452--40474},
  year={2023},
  organization={PMLR}
}

@article{amortila2025reinforcement,
  title={Reinforcement Learning under Latent Dynamics: Toward Statistical and Algorithmic Modularity},
  author={Amortila, Philip and Foster, Dylan J and Jiang, Nan and Krishnamurthy, Akshay and Mhammedi, Zak},
  journal={Advances in Neural Information Processing Systems},
  volume={37},
  pages={133007--133091},
  year={2025}
}

@inproceedings{wang2024deep,
  title={Deep reinforcement learning for early diagnosis of lung cancer},
  author={Wang, Yifan and Zhang, Qining and Ying, Lei and Zhou, Chuan},
  booktitle={Proceedings of the AAAI Conference on Artificial Intelligence},
  volume={38},
  number={20},
  pages={22410--22419},
  year={2024}
}

@article{radosavovic2024real,
  title={Real-world humanoid locomotion with reinforcement learning},
  author={Radosavovic, Ilija and Xiao, Tete and Zhang, Bike and Darrell, Trevor and Malik, Jitendra and Sreenath, Koushil},
  journal={Science Robotics},
  volume={9},
  number={89},
  pages={eadi9579},
  year={2024},
  
}

@article{oord2018representation,
  title={Representation learning with contrastive predictive coding},
  author={Oord, Aaron van den and Li, Yazhe and Vinyals, Oriol},
  journal={arXiv preprint arXiv:1807.03748},
  year={2018}
}

@article{zhuang2025tdmpbc,
  title={TDMPBC: Self-Imitative Reinforcement Learning for Humanoid Robot Control},
  author={Zhuang, Zifeng and Shi, Diyuan and Suo, Runze and He, Xiao and Zhang, Hongyin and Wang, Ting and Lyu, Shangke and Wang, Donglin},
  journal={arXiv preprint arXiv:2502.17322},
  year={2025}
}

@inproceedings{wang2024negatively,
  title={Negatively correlated ensemble reinforcement learning for online diverse game level generation},
  author={Wang, Ziqi and Hu, Chengpeng and Liu, Jialin and Yao, Xin},
  booktitle={The Twelfth International Conference on Learning Representations},
  year={2024}
}

@inproceedings{zheng2024effective,
  title={Effective Representation Learning is More Effective in Reinforcement Learning than You Think},
  author={Zheng, Jiawei and Song, Yonghong},
  booktitle={2024 IEEE International Conference on Robotics and Automation (ICRA)},
  pages={9176--9182},
  year={2024},
  organization={IEEE}
}

@inproceedings{mao2025offline,
  title={Offline reinforcement learning with ood state correction and ood action suppression},
  author={Mao, Yixiu and Wang, Qi and Chen, Chen and Qu, Yun and Ji, Xiangyang},
  booktitle={Advances in Neural Information Processing Systems},
  volume={37},
  pages={93568--93601},
  year={2025}
}

@article{fan2025solve,
  title={How to Solve Contextual Goal-Oriented Problems with Offline Datasets?},
  author={Fan, Ying and Li, Jingling and Swaminathan, Adith and Modi, Aditya and Cheng, Ching-An},
  journal={Advances in Neural Information Processing Systems},
  volume={37},
  pages={99433--99463},
  year={2025}
}

@inproceedings{chemingui2024offline,
  title={Offline model-based optimization via policy-guided gradient search},
  author={Chemingui, Yassine and Deshwal, Aryan and Hoang, Trong Nghia and Doppa, Janardhan Rao},
  booktitle={Proceedings of the AAAI Conference on Artificial Intelligence},
  volume={38},
  number={10},
  pages={11230--11239},
  year={2024}
}

@article{luo2024dtr,
  title={DTR-bench: an in silico environment and benchmark platform for reinforcement learning based dynamic treatment regime},
  author={Luo, Zhiyao and Zhu, Mingcheng and Liu, Fenglin and Li, Jiali and Pan, Yangchen and Zhou, Jiandong and Zhu, Tingting},
  journal={arXiv preprint arXiv:2405.18610},
  year={2024}
}

@inproceedings{yu2024cheaper,
  title={Cheaper and faster: Distributed deep reinforcement learning with serverless computing},
  author={Yu, Hanfei and Li, Jian and Hua, Yang and Yuan, Xu and Wang, Hao},
  booktitle={Proceedings of the AAAI Conference on Artificial Intelligence},
  volume={38},
  number={15},
  pages={16539--16547},
  year={2024}
}

@article{luo2025optimistic,
  title={Optimistic Critic Reconstruction and Constrained Fine-Tuning for General Offline-to-Online RL},
  author={Luo, Qin-Wen and Xie, Ming-Kun and Wang, Yewen and Huang, Sheng-Jun},
  journal={Advances in Neural Information Processing Systems},
  volume={37},
  pages={108167--108207},
  year={2025}
}

@article{shaheen2025reinforcement,
  title={Reinforcement Learning in Strategy-Based and Atari Games: A Review of Google DeepMinds Innovations},
  author={Shaheen, Abdelrhman and Badr, Anas and Abohendy, Ali and Alsaadawy, Hatem and Alsayad, Nadine},
  journal={arXiv preprint arXiv:2502.10303},
  year={2025}
}

@article{liu2024adaptive,
  title={Adaptive pessimism via target Q-value for offline reinforcement learning},
  author={Liu, Jie and Zhang, Yinmin and Li, Chuming and Yang, Yaodong and Liu, Yu and Ouyang, Wanli},
  journal={Neural Networks},
  volume={180},
  pages={106588},
  year={2024},
  publisher={Elsevier}
}

@article{qi2023adaptive,
  title={An adaptive reinforcement learning-based multimodal data fusion framework for human--robot confrontation gaming},
  author={Qi, Wen and Fan, Haoyu and Karimi, Hamid Reza and Su, Hang},
  journal={Neural Networks},
  volume={164},
  pages={489--496},
  year={2023},
  publisher={Elsevier}
}

@article{liu2024segmenting,
  title={Segmenting medical images with limited data},
  author={Liu, Zhaoshan and Lv, Qiujie and Lee, Chau Hung and Shen, Lei},
  journal={Neural Networks},
  volume={177},
  pages={106367},
  year={2024},
  publisher={Elsevier}
}
\end{document}